\documentclass[10pt,letterpaper]{article}
\usepackage[dvipsnames]{xcolor}

\usepackage[top=0.85in,left=2.75in,footskip=0.75in]{geometry}

\usepackage{comment}

\usepackage{amsmath,amssymb}

\usepackage{changepage}

\usepackage{textcomp,marvosym}

\usepackage{cite}

\usepackage{nameref,hyperref}

\usepackage[right]{lineno}

\usepackage[nopatch=eqnum]{microtype}
\DisableLigatures[f]{encoding = *, family = * }

\usepackage{array}

\usepackage{soul}

\newcolumntype{+}{!{\vrule width 2pt}}

\newlength\savedwidth

\raggedright
\usepackage[aboveskip=1pt,labelfont=bf,labelsep=period,justification=raggedright,singlelinecheck=off]{caption}

\makeatletter
\renewcommand{\@biblabel}[1]{\quad#1.}
\makeatother

\usepackage{lastpage,fancyhdr,graphicx}
\usepackage{epstopdf}
\fancyheadoffset[L]{2.25in}
\fancyfootoffset[L]{2.25in}
\usepackage{tabularx}

\begin{document}
\vspace*{0.2in}

\begin{flushleft}
{\Large
\textbf\newline{Sustainable Visions: Unsupervised Machine Learning Insights on Global Development Goals} 
}
\newline
\\

Alberto Garc\'ia-Rodr\'iguez\textsuperscript{1,7,8},
Matias Núñez\textsuperscript{2,3},
Miguel Robles P\'erez\textsuperscript{4},
Tzipe Govezensky\textsuperscript{5},
Rafael A. Barrio\textsuperscript{1*},
Carlos Gershenson\textsuperscript{6,8,9,10},
Kimmo K.Kaski\textsuperscript{7},
Julia Tagüeña\textsuperscript{4,8},
\\
\bigskip
\textbf{1} Instituto de Física, Universidad Nacional Autónoma de México, Coyoac\'an, Ciudad de M\'exico, M\'exico.
\\
\textbf{2} Consejo Nacional de Investigaciones Científicas y Técnicas, Buenos Aires, Argentina
\\
\textbf{3} Instituto de Investigaciones en Biodiversidad y Medioambiente, Universidad Nacional del Comahue, Bariloche, Argentina
\\
\textbf{4} Instituto de Energ\'ias Renovables, Universidad Nacional Aut\'onoma de M\'exico, Temixco, Morelos, M\'exico
\\
\textbf{5} Instituto de Investigaciones Biomédicas, Universidad Nacional Autónoma de México, Coyoac\'an, Ciudad de M\'exico, M\'exico.
\\
\textbf{6} Department of Systems Science and Industrial Engineering Thomas J. Watson College of Engineering and Applied Science, State University of New York at Binghamton, Binghamton, New York, USA.
\\
\textbf{7} Department of Computer Science, Aalto University School of Science, Espoo, Finland
\\
\textbf{8} Centro de Ciencias de la Complejidad, Universidad Nacional Autónoma de México, Coyoac\'an, Ciudad de M\'exico, M\'exico.
\\
\textbf{9} Instituto de Investigaciones en Matem\'aticas Aplicadas y en Sistemas, Universidad Nacional Aut\'onoma de M\'exico, Coyoac\'an, Ciudad de M\'exico, M\'exico.
\\
\textbf{10} Lakeside Labs GmbH, Klagenfurt am Wörthersee, Austria

\bigskip

%
%





* barrio@fisica.unam.mx 

\end{flushleft}
\section*{Abstract}
The 
2030 Agenda for Sustainable Development of the United Nations outlines 17 goals for countries of the world to address global challenges in their development. 
However, the progress of countries towards these goal has been slower than expected and, consequently, there is a need to investigate the reasons behind this fact.
In this study, we have used a novel data-driven methodology to analyze time-series data for over 20 years (2000-2022) from 107 countries 
using unsupervised machine learning (ML) techniques. 
 Our analysis reveals strong positive and negative correlations between certain SDGs (Sustainable Development Goals).
 Our findings show that progress toward the SDGs is heavily influenced by geographical, cultural and socioeconomic factors, with no country on track to achieve all the goals by 2030. This highlights the need for a region-specific, systemic approach to sustainable development that acknowledges the complex interdependencies between 
 the goals and the variable 
 capacities of countries to reach them.  
For this our machine learning based approach 
 provides a robust framework for developing efficient and 
 data-informed strategies to promote cooperative and targeted initiatives for sustainable progress.


\section{Introduction}

In 2015, the 2030 Agenda for Sustainable Development Goals (SDG) of the United Nations \cite{URL:Goals} was established for countries of the world to follow 17 SDGs, 169 targets and 231 globally comparable indicators in their development process (for details, see the {\bf{Supplementary material S1}} section). Since then, the world has faced several setbacks, such as the COVID-19 pandemic, acceleration of climate change, and serious armed conflicts that have slowed down this process. For these reasons and because the SDGs are
variably interlinked or interrelated with each other, it seems difficult or even unlikely for countries to achieve the SDGs by 2030. Furthermore, in 2023 an independent group of scientists appointed by the UN to assess the progress of the world in achieving the SDGs and to recommend how to move forward, stated~\cite{ShirinMalekpour2023}: 
“The world is not on track to achieve any of the 17 SDGs and cannot rely on change to happen organically. Without accelerated action, the ambitious plan that the world signed up to in 2015 will fail."

Therefore, it seems obvious that we lack understanding of the complex inter-linkages between different SDGs and what are the drivers leading to their changes. This in turn points out the need for more research, which is made possible by the vast amount of data that the 231 global indicators of the 17 SDGs have produced so far. A thorough analysis of these data is likely to give us some insight into the correlations between, e.g., armed conflicts, climate change, social inequalities, global crime, etc. 
This would serve as a necessary precursor for the next steps towards sustainability.

As direct interpretation of the available raw data and because the roles and interlinkages between SDGs are difficult to decipher, our aim in this study is to get deeper understanding and insight into them using a modern data analysis approach. We will do this by using unsupervised machine learning methods as they can reveal groups of countries with similar behavior as regards to 
the 17 SDG's scores. For example, in the case of diversity, 
even subtle differences between countries otherwise quite similar should be considered when proposing a new policy agenda. All 
this should provide valuable information to decision makers at the national and local levels to make well-informed policy choices for achieving 
the goals and aligning 
with the 2030 Agenda. The best practices for decision making involve identifying challenges, using data to form insight, 
and to employ 
risk analysis while taking into account environmental constraints and local conditions \cite{Panpatte2019}.

Sustainability, like many other human endeavors, can be represented as a dynamic and complex network with nonlinear interactions and interdependencies between the Sustainable Development Goals~\cite{ComplexityExplained}. Therefore, as the SDGs are interconnected and influenced by the intricate web of environmental, political, economic, cultural and social factors, they should not be considered in isolation. 
Recent progress in complex systems research has enhanced our ability to analyze sustainability using new systemic approaches~\cite{Fisher2024}. Here we will do this by employing the network science approach and its 
tools, while recognizing that cultural factors play an important role and are difficult to isolate within the global indicators.

So far several studies with different degree of coverage and thoroughness have been conducted to analyze the interlinkages between the SDGs~\cite{Bennich2020,Urrutia,Pradhan2017,CLING2022,BALISWAIN2021}. Among them {Ref. \cite{Bennich2020} constitutes 
a comprehensive review of 70 previous papers, however, as it is dated for 2020 it does not cover the effect of such recent events as COVID-19. The focus in ref. \cite{Urrutia} is very local using only Mexican data.} 
In a number of other studies various methodological approaches have been used,  e.g., 
the nonparametric Spearman’s rank correlation analysis for assessing 
monotonic relationships between all possible combinations of the indicator data pairs \cite{Pradhan2017}, the Multiple Factor Analysis (MFA) used to measure the correlation between indicators \cite{CLING2022}, and the network analysis approach \cite{BALISWAIN2021}. In contrast to these methods we will in our study use a 
combination of techniques
to study in detail these interlinkages,
which can be either positive or negative, indicating that progress on one goal can enhance or hamper the progress on other goals. For a country, identifying these synergies and trade-offs before setting its policy priorities is crucial to avoid creating new problems while solving existing ones. This interesting issue of SDG interlinkings has been examined in a recent review
study~\cite{Bennich2023}, which concluded that our understanding of sustainability is still quite rudimentary. 

In our study, we will use an unsupervised machine learning multimethod approach with three objectives in mind: i) to examine the changes over time in the 17 SDGs for all countries, identifying possible classifications and discovering similarities and differences in their patterns; ii) to understand how the SDGs correlate 
across different countries; and iii) to investigate the collective dynamics of similar countries to measure the rate of progress towards achieving 
the SDGs has been generated in the 21$^{\mathrm{st}}$century. Our methods uncover distinct SDG patterns associated with well-defined clusters of countries, each showing different dynamical behaviors towards ideal scores.

This paper is organised as follows: a detailed explanation of the methodology is to be founds in Section 2. Section 3 presents the results, including multimethod machine learning results, followed by an analysis of the correlations between the SDGs, and the dynamics within the clusters and finally, Sections 4 we discuss our findings and in Section 5 there are some important concluding remarks. 

\section{Materials and methods}

\begin{figure}[ht!]
\centering
\includegraphics[width=1\linewidth]{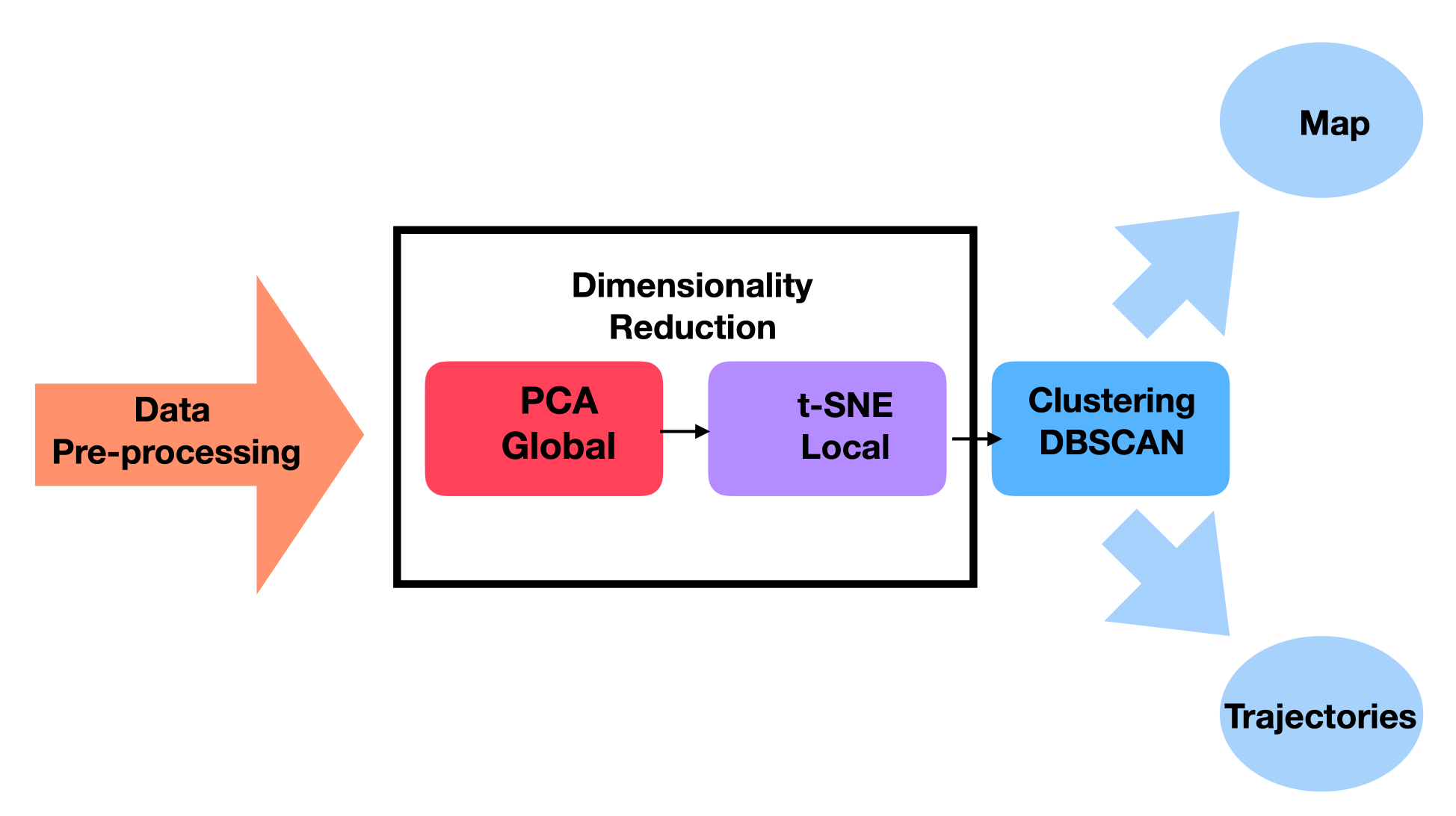} 
\caption{{\bf A three stage unsupervised learning pipeline for data analysis}.
The first stage involves data preprocessing, where the dataset undergoes standardization, normalization, and cleaning procedures to address missing values and noise. The second stage consists of sequential dimensionality reduction by PCA to identify principal components to capture global data structure, followed 
by t-SNE which preserves local relationships in the reduced dimensional space. The final stage applies a clustering algorithm (DBSCAN) to identify distinct groups within the processed data. Countries in each cluster are mapped and the mean trajectories towards ideal scores are calculated. Arrows between the stages indicate the sequential flow of data through the pipeline, with the output of each stage serving as the input for the subsequent stage.}
\label{fig:flowchart}
\end{figure}

In order to explore, analyse, and understand the nature of interdependences in the complex dataset of SDGs we have devised a novel three-stage unsupervised learning pipeline, depicted in Fig. \ref{fig:flowchart}. 
It consist of dimensionality reduction and clustering by which we are 
able to effectively simplify and analyze the high-dimensional data, thus making hidden patterns more evident. The use of Principal Component Analysis (PCA) ~\cite{pearson1901liii,Jollife2016} helps us to reduce the complexity of the data by identifying the main 
trends and providing a global overview, while t-Distributed Stochastic Neighbor Embedding (t-SNE)~\cite{JMLR:v9:vandermaaten08a,nunez2019exploring} focuses on preserving local relationships, thus allowing for a more detailed exploration of similarities between countries. Finally, the DBSCAN clustering technique~\cite{DBSCAN} helps to identify distinct groups, without prior assumptions about the number or shape of clusters, making it suitable for real data, as well as for artificial data. We believe 
that this combination of analysis methods 
is the most suitable in providing a global insight into  the information  embedded in the complex datasets.

\subsection{Data}

The data in this study was collected from the “Online database for the Sustainable Development Report 2023”\cite{sachs20232023}, including only countries without missing data. The original data include 166 countries, but 15 of them lack information on Goal 1 (no poverty), 17 countries did not have information about Goal 10 (reduce inequalities), and 40 countries that lack coastal regions do not have scores for Goal 14 (life below water). Therefore, only 107 countries were included in our analysis.

In this study we used also data of Gross Domestic Product per capita (GDP/c) of every country for the 2022 collected  from “Worldometer” \cite{worldometers}. This data serves as a measure of each country's economic activity as well as overall capability to achieve SDGs. 

\subsection{Principal Component Analysis (PCA).}

Principal component analysis (PCA)\cite{pearson1901liii,Jollife2016} is widely used as a dimensionality reduction technique. It is often employed to visualize multidimensional data in a two dimensional plane. Taking into account that, if separate groups exist in the original data, it will be possible to see them when a linear projection is made into the plane that explains most of the variability in the dataset. This is achieved by determining the first axis in the direction with maximal variability, the second axis, perpendicular to the first one, is determined in such a way that it exhibits the maximum 
variance, not exhibited by the first axis. These two axes are known as PC1 and PC2, respectively. The process can be repeated to find a third axis, a fourth one, and so on. The coordinates of each point in the PCA plane are linear combinations of all the original variables.

PCA was performed using standardized data, that is, the mean and standard deviations were calculated for each SDG throughout the years 2000 to 2022. The corresponding mean was then subtracted from each observation, and the result was divided by the SDG's standard deviation.

\subsection{t-Distributed Stochastic Neighbor Embedding (t-SNE) and clustering.}

The t-SNE~\cite{JMLR:v9:vandermaaten08a,nunez2019exploring} algorithm is a powerful technique for visualizing high-dimensional data by mapping it into a lower-dimensional space, typically into 2 or 3 dimensions. The primary objective of t-SNE is to represent the underlying structure of the data by placing similar data points close together and dissimilar points further apart in the lower-dimensional space. This is achieved by converting the pairwise Euclidean distances between data points into conditional probabilities, which reflect the likelihood of points being neighbors in the original high-dimensional space. The resulting visualization preserves the local  relationships within the data, making it easier to identify patterns and clusters. A more in depth explanation of the method can be found in {\bf{Supplementary material S1}} section. 

To further analyze the structure revealed by t-SNE, we employed the DBSCAN~\cite{DBSCAN} clustering technique, a non-parametric algorithm that groups closely packed points into clusters. Unlike traditional clustering methods, DBSCAN does not require the number of clusters to be specified in advance and can effectively identify clusters of varying shapes and sizes. Points that do not belong to any cluster, often located in low-density regions, are marked as outliers, providing additional insights into the data distribution.

In our analysis, we applied the t-SNE dimensionality reduction technique followed by the DBSCAN clustering algorithm, carefully tuning their key parameters to ensure optimal results. For t-SNE, the primary parameter is perplexity, which influences the number of effective neighbors for each point and balances the local and global aspects of the data. We selected a perplexity value of 50 after conducting a grid search across a range of values (5 to 50). This value turned out to provide the best representation for our dataset by effectively capturing local relationships between countries while preserving broader trends in their Sustainable Development Goal (SDG) performance. The chosen perplexity was visually validated by examining the resulting 2D t-SNE plots, where the clusters were well separated and meaningful. We also experimented with higher-dimensional t-SNE embeddings and found that similar clustering patterns emerged, indicating the robustness of our approach.
Once the t-SNE dimensionality reduction was performed, we used DBSCAN for clustering directly in the 2D t-SNE plane. The main parameter in DBSCAN, epsilon, defines the neighborhood radius within which points are considered part of the same cluster. We tested various values for epsilon and selected the one that resulted in well-defined clusters with minimal noise points, ensuring that the clusters captured meaningful patterns in the data. The chosen epsilon parameter was correlated visually with the 2D t-SNE plots to identify distinct groups of countries based on their SDG trajectories. Similar clustering results were also observed when using higher-dimensional t-SNE projections, confirming the stability of our clustering approach across different dimensionalities. By carefully tuning these parameters, we achieved a balance between local structure and global separations, resulting in insightful and robust clusters that reveal distinct SDG progress patterns among countries.

\section{Results}

\subsection{Global view of  SDG scores}

\begin{figure}[ht!]
\centering
\includegraphics[width=1\linewidth]{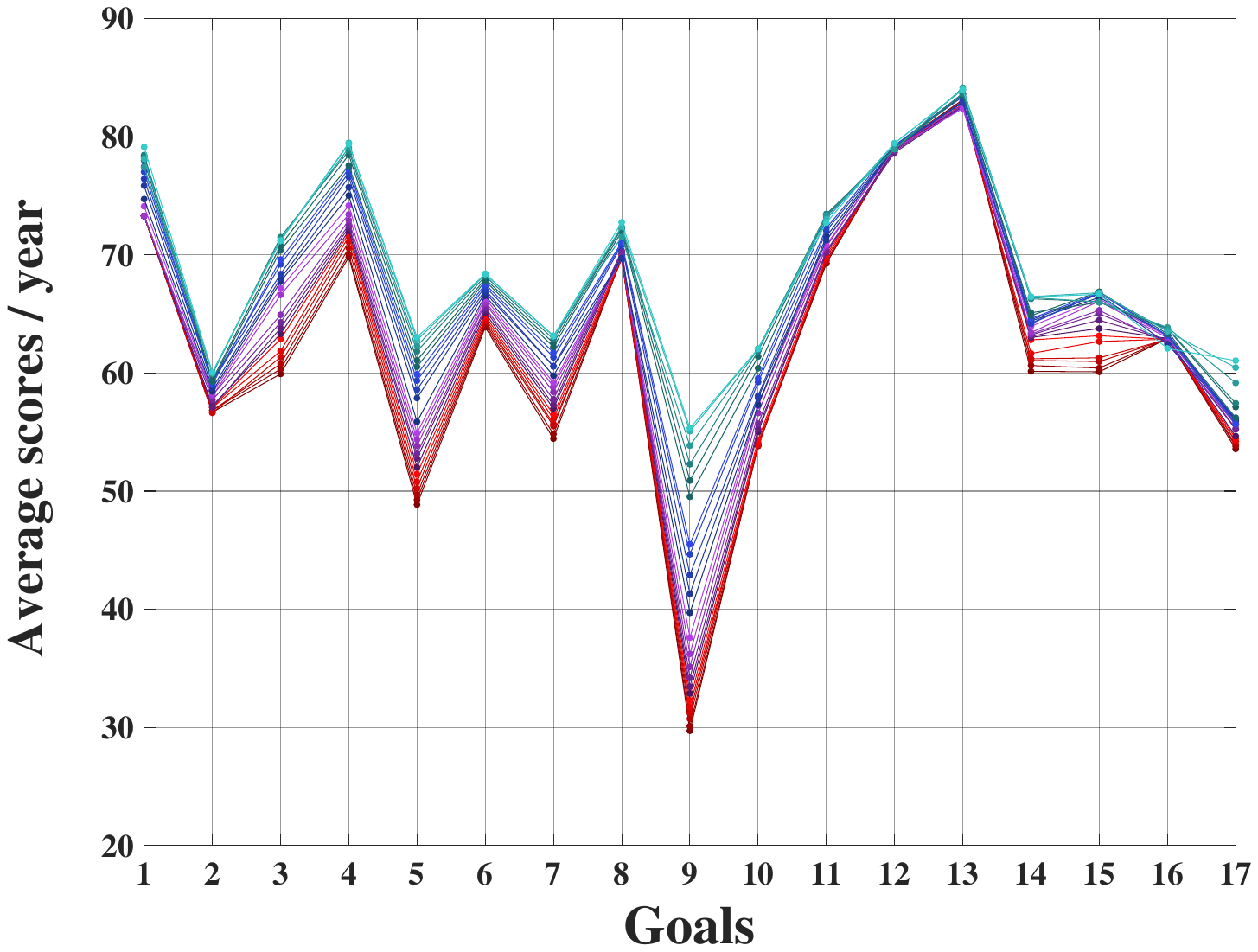}
\caption{{\bf Parallel plot of the 17 SDG.}\\ Comparison among all the yearly average SDG scores, from 2000 to 2022. Scale color goes from dark red (2000) to light blue (2022).}
\label{fig:ParPlot}
\end{figure}

Given that the effectiveness of sustainability relies on global collaboration, our initial findings focus on the average Sustainable Development Goals (SDGs) scores at a global scale. Figure ~\ref{fig:ParPlot} illustrates the yearly mean scores for each goal, enabling simultaneous observation of the values of all 17 goals over the study period. Notably, Goal 9 (Industry, innovation, and infrastructure) had the lowest observed average score (29.7 in the year 2000), but it demonstrated a consistent increase over time, reaching an average of 55.3 such that, in 2022, all average scores were above 50. Conversely, goals 12, 13, and 16 (responsible production and consumption; climate action; and peace, justice, and strong institutions) did not exhibit significant changes over the years. This result may be attributed to socio-economic circumstances. 

\subsection{Principal Component analysis}

Due to the potential interactions among the Sustainable Development Goals (SDGs), a multivariate analysis is required for the examination of the SDGs. Likewise, average behaviors of all the countries, grouped based on similar scores, will provide further understanding about achievements in sustainability goals. At first, we utilize principal component analysis (PCA) because it is straightforward to interpret the results as a linear projection of the original data.

\begin{figure}[ht]
\centering
\includegraphics[width=1\linewidth]{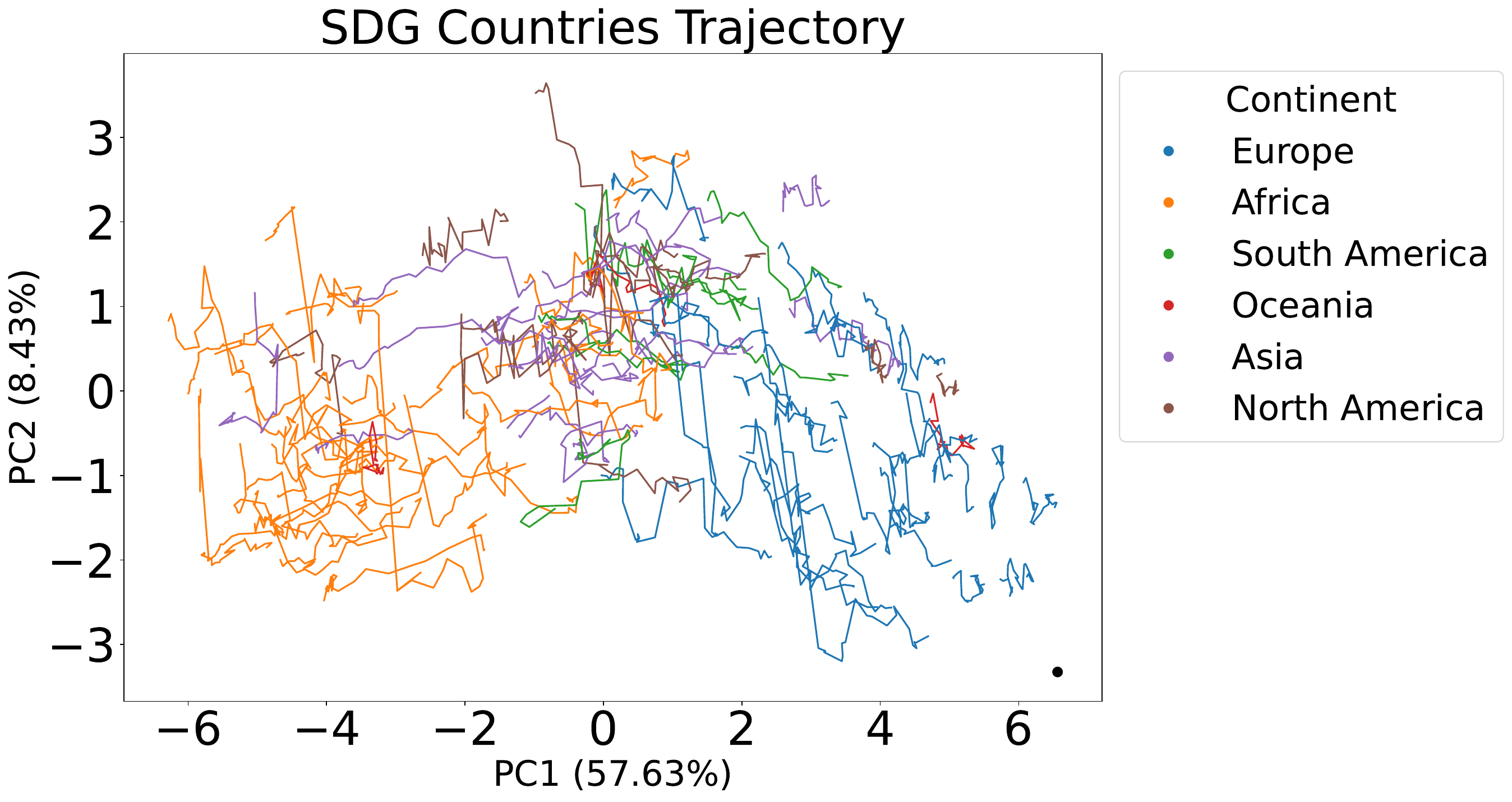}
\caption{{\bf Principal Component Analysis (PCA) of Global SDG Progress from 2000–2022.}\\ This two-dimensional plot represents the positioning of countries concerning the Sustainable Development Goals. Lines link between points in different years for a country. Oscillating paths suggest varying rates of advancement. The black dot in the lower right corner represents the ideal goal for 2030.}
\label{fig:GOALS_PCA}
\end{figure}

In our analysis,
Figure~\ref{fig:GOALS_PCA} presents a reductionist perspective by utilizing PCA to simplify the high-dimensional SDG data into a  two-dimensional plane. The plot signifies the progression of countries with respect to the 17 United Nations Sustainable Development Goals (SDGs) from 2000 to 2022. The lines connecting the data points for each country depict the trajectory of that country's performance or alignment with the SDGs over the years.
Oscillating paths may indicate periods of both advances and setbacks, while short trajectories correspond to countries that remained relatively stable in their SDG-related metrics over the years.

Our analysis shows that the primary principal component (PC1) explains 57.5\% of the aggregate variance, highlighting its important role in capturing primary patterns in the data. In contrast, the second principal component (PC2) is responsible for 8.7\% of the variance, indicating that this representation preserves up to 66.2\% of the total variance. There is no discernible separation of clusters. Only when the PCA countries are color-coded by continent is it possible to see some clustering, showing that SDGs scores are similar between countries on the same continent. Incorporating information from additional components would improve separation, but at the expense of visual complexity.

\begin{figure}[ht!]
\centering
\includegraphics[width=1\linewidth]{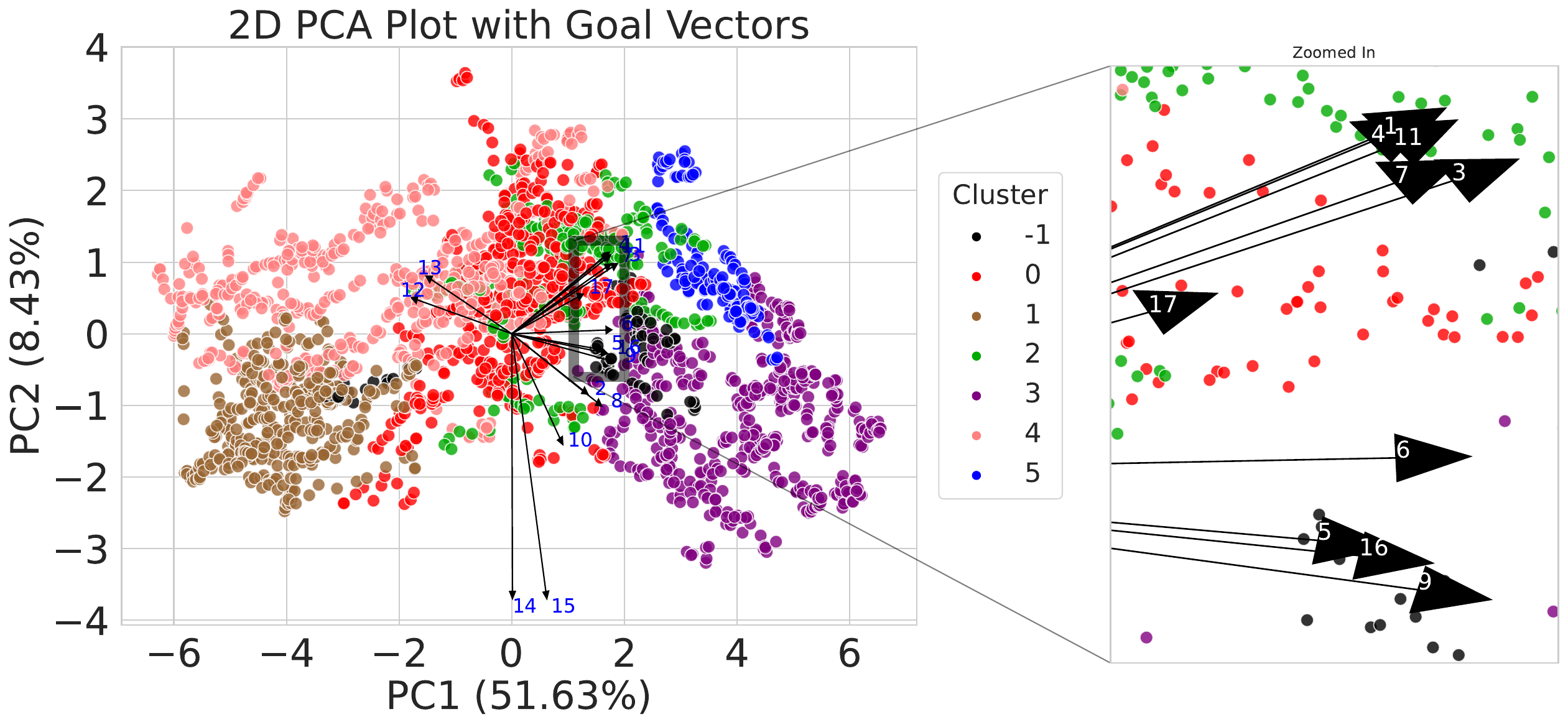}
\caption{{\bf PCA with SDG Vectors.}\\ An enhanced view of the PCA plot introduces vectors representing each of the 17 Sustainable Development Goals. The orientation and magnitude of these vectors offer insights into the influence and correlation of individual goals within the principal components.}
\label{fig:GOALS_PCA_VEC}
\end{figure}

An alternative version of the PCA plot is illustrated in Figure~\ref{fig:GOALS_PCA_VEC}, which improves comprehension by incorporating SDG vectors. 
Projections of the original 17 SDGs onto the reduced two-dimensional space are represented by the vectors superimposed on the PCA plot. The orientation and magnitude of these vectors provide an engaging visual representation, allowing for an understanding of how each SDG correlates with the primary components of variation, PC1 and PC2.

Each vector points in the direction of the maximum increase for a particular SDG, and its magnitude (or length) gives an indication of the importance of that goal in the variability captured by the two principal components. In other words, longer vectors suggest that the respective SDG has a strong influence on the dataset's variance structure, while shorter vectors might indicate a lesser contribution.

PC1 divides countries with high scores in SDGs 12 and 13 to the left (primarily African countries) and countries with high scores when summing several SDGs scores, including 3, 6, 7, 9, and 11, to the right (primarily European countries). PC2 separates countries mainly according to the scores of SDGs 14, 15, and 10, with high scores at the bottom and low scores at the top.

It is important to note that goals with vectors that are closely aligned or point in a similar direction have positive correlations, as illustrated below (Figure~\ref{fig:GOALS_PCA_VEC}, see for example vectors corresponding to SDG's 2 and 8 or SDG's 5,16 and 9). This suggests that nations that exhibit success in one of these objectives are likely to succeed in the others. In contrast, goals with vectors pointing in opposite directions, such as vectors 13 and 8, exhibit negative correlations. Additionally, where two vectors are nearly orthogonal (i.e., they form a right angle, as in vectors 14 and 16), this suggests that there is minimal to no correlation between their respective objectives in the current dataset.

The spatial configuration of the vectors in relation to the data points on the scatter plot reveals additional insights. Sustainable Development Goals (SDGs) with vectors pointing towards regions with a higher concentration of countries may indicate that these objectives are particularly pertinent or prioritized by those nations. Conversely, SDGs with vectors oriented towards regions with lower population density may suggest challenges or diminished emphasis on those specific goals among the countries represented.

\subsection{Clustering Analysis: Diverse SDG Patterns for Different Geographical Regions.}

While the PCA plot provides a linear projection that captures the variance and relationships within the data, t-SNE (t-distributed stochastic neighbor embedding) offers a finer, non-linear perspective. By focusing on preserving the local structure and emphasizing the subtle similarities and differences between data points, t-SNE reveals patterns and relationships that may be obscured in linear methods, thus providing a more comprehensive understanding of the underlying data structure.

\begin{figure}[ht]
\centering
\includegraphics[width=1\linewidth]{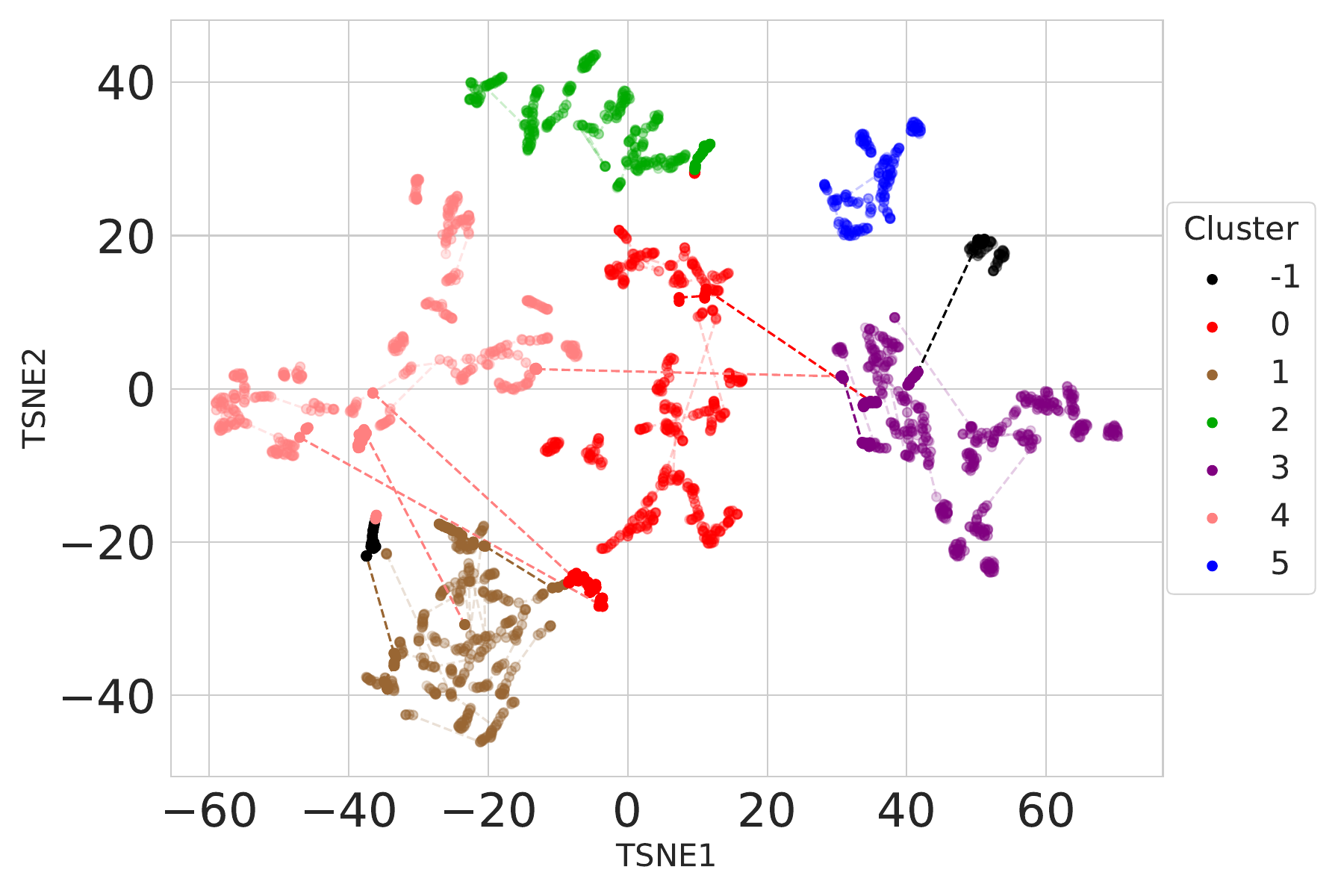}
\caption{{\bf t-SNE Visualization with DBSCAN Clustering.}\\ A detailed t-SNE plot colored by clusters determined using the DBSCAN algorithm. We use transparency for most dots to reduce visual clutter, highlighting only those that change between clusters with full color. These changes are further emphasized by connecting the highlighted dots with dotted lines.   The countries in each clusters are enumerated in S1 Table 2.1 (in {\bf{Supplementary material S1}} section)}
\label{fig:GOALS_TSNE_clustering}
\end{figure}

We employed both methods in tandem to analyze SDG data from 2000 to 2022. Specifically, we provided t-SNE with the 10 principal components derived from PCA, which together account for 95\% of the variance, we use 50 as perplexity input (this term is explained in  {\bf{Supplementary material S1}} section) . Subsequently, we applied the DBSCAN algorithm to identify clusters within the reduced t-SNE space. This approach yielded more distinct and cohesive clusters among regions with similar SDG patterns, while also highlighting the  separations between regions with  different patterns (Figure~\ref{fig:GOALS_TSNE_clustering}).

\begin{figure}[ht]
\centering
\includegraphics[width=1\linewidth]{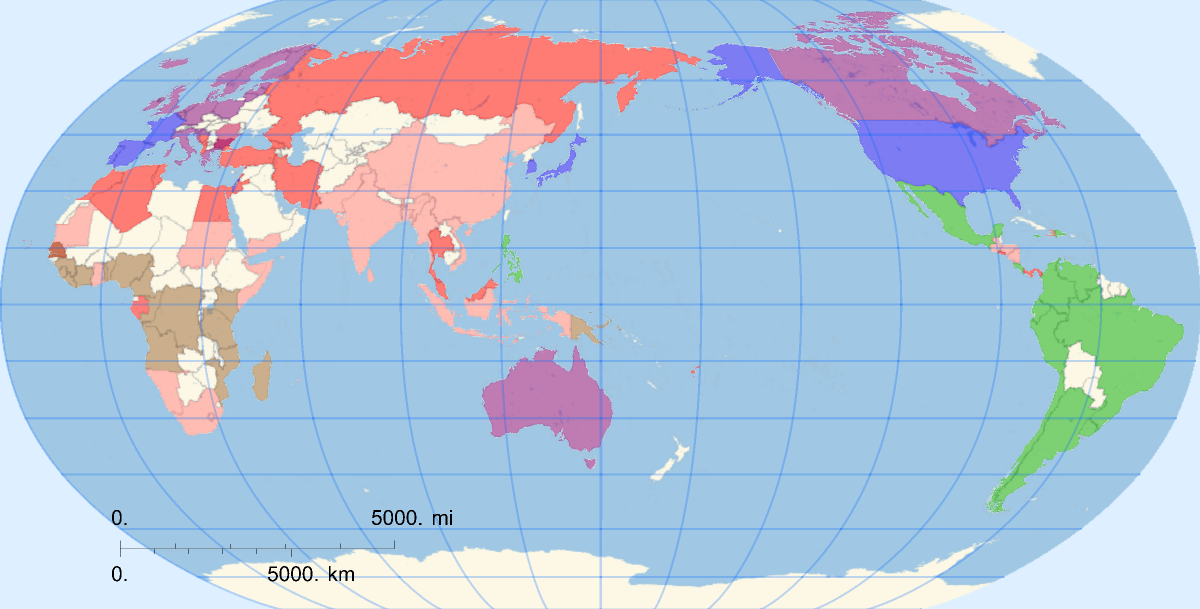}
\caption{{\bf Colored countries map.}\\ Colored map from countries in S1 Table 2.1 (in {\bf{Supplementary material S1}} section). Cluster 0 is in red, cluster 1 brown, cluster 2 green, cluster 3 purple, cluster 4 light red, cluster 5 blue.}
\label{fig:Map}
\end{figure}

This exploration allows for a granular exploration of SDG data from 2000 to 2022. A deeper view  is only possible when we see the clusters in a world map (see Figure~\ref{fig:Map}) . This visualization provides a detailed view of how different regions or continents have evolved concerning the UN SDGs. As a result, we see tighter clusters for regions with similar SDG patterns and more significant separations between regions with different patterns.

In general, each country remained within the same cluster throughout the period 2000-2022. However, ten countries exhibited trajectories that spanned across two distinct clusters (see dotted lines in Figure~\ref{fig:GOALS_TSNE_clustering}). This shift was primarily driven by improvements in their performance on specific SDGs; notably, eight of these countries showed significant progress in SDG 8 (decent work and economic growth). It is important to note that the number $-1$ does not refer to a cluster, as explained in Appendix B.

Countries grouped within each cluster are presented in S1 Table 2.1 (in {\bf{Supplementary material S1}} section). Interestingly, the clusters are primarily organized by geographical proximity (see Figure~\ref{fig:Map}), though other influencing factors may include the development level of the countries and existing commercial agreements between them. Cluster 0 predominantly consists of countries from Asia and Africa, with some representation from the Americas, Europe, and Oceania. In contrast, Clusters 1, 2, and 3 are more homogeneous, comprising African, American, and European countries, respectively. Cluster 4 includes a mix of African and Asian countries, while Cluster 5, the smallest, consists of seven developed countries.

\begin{figure}[ht]
\center
\includegraphics[width=1\linewidth]{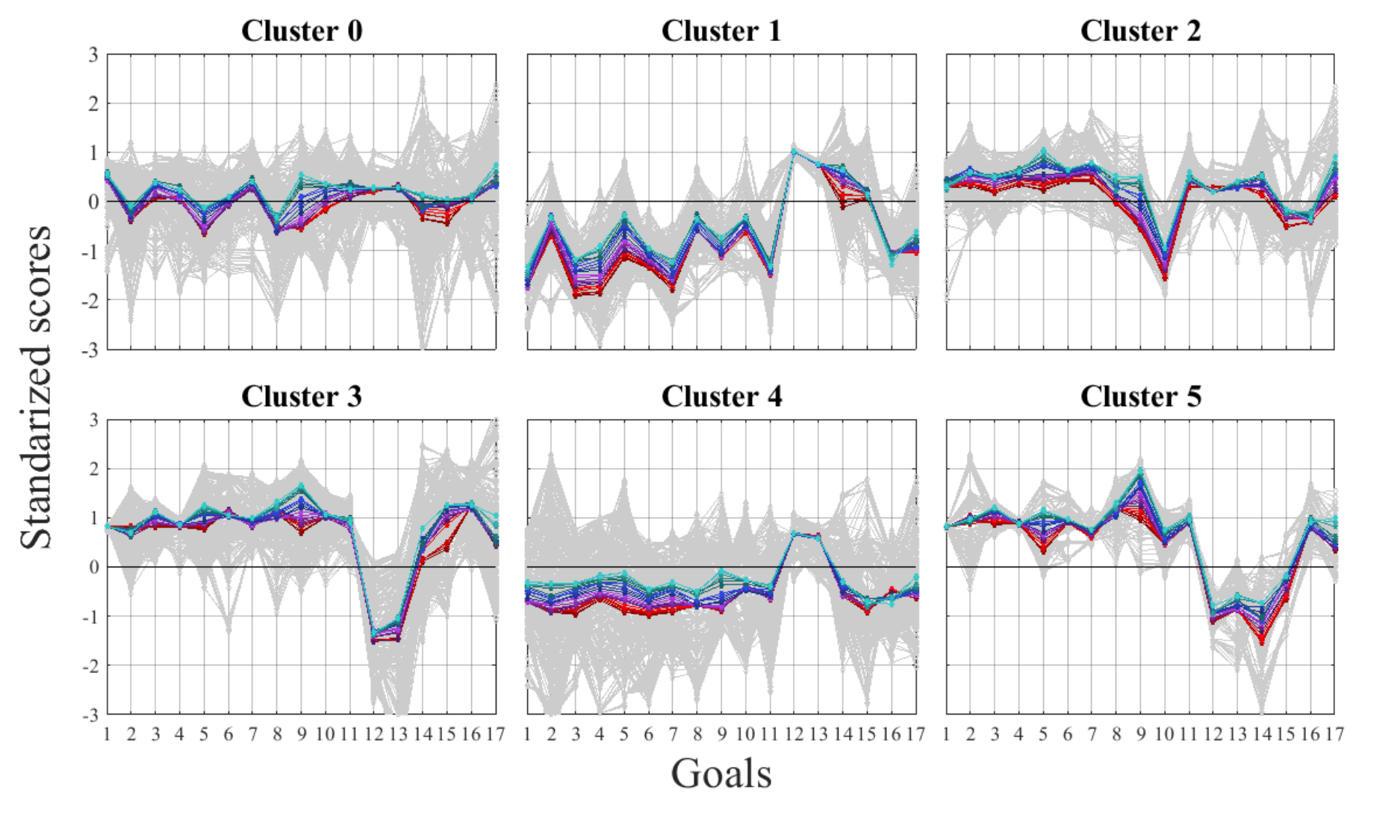} 
\caption{{\bf SDG patterns observed in different clusters.} \\Data are standardized by subtracting cluster’s average for each SDG and dividing by the corresponding standard deviation; therefore, values above the average are represented as positive, and values below the average are negative. Gray lines are standardized SDGs of the countries in the cluster, colored lines are the yearly average of standardized SDG scores, colors are the same as in Fig.\ref{fig:ParPlot}: reds 2000 to blues 2022}
\label{fig:ClutSNE}
\end{figure}

Clustering allowed us to identify distinct regional patterns concerning the United Nations SDGs from 2000 to 2022 (see Figure~\ref{fig:ClutSNE}). In this figure standardized SDG scores are presented (as described in the Methods, PCA section), where a value of zero on the y-axis corresponds to the mean SDG value across the dataset. Scores above zero indicate performance above the average for a given SDG, while scores below zero reflect performance below the average.
Cluster zero is particularly heterogeneous, encompassing countries from all continents with diverse SDG patterns (as shown by the variability of the gray lines in  Figure~\ref{fig:ClutSNE}), resulting in average standardized SDG scores that fluctuate around the mean. Other clusters display greater homogeneity. In Cluster 1, the annual averages for most SDGs are negative, with the exceptions of SDG 12 (responsible production and consumption), SDG 13 (climate action), SDG 14 (life below water), and SDG 15 (life on land). Cluster two shows predominantly positive annual averages for the SDGs, except for SDG 10 (reduced inequalities), SDG 15, and SDG 16 (peace, justice, and strong institutions). In Cluster 3, all SDG annual averages exceed the mean, except for SDG 12 and SDG 13, which contrasts with Cluster 4, where the annual scores are below the mean for all SDGs except SDGs 12 and 13. Cluster 5 is similar to Cluster 3, but in addition to SDGs 12 and 13, SDGs 14 and 15 are also below the average.

We note that the Clusters 
differ not only in geographical localization and SDG patterns but also in their average GDP/c (Gross Domestic Product per capita), a widely used economic indicator. Among the included countries, those belonging to Cluster 1 have the lowest average GDP/c (\$1,464.44 $\pm$\$883.72). 
The average GDP/c value for the countries in Cluster 4 is \$3,941.89 $\pm$ \$3,113.04 while for the 
countries grouped in Cluster 0 it is \$7,111.55 $\pm$ \$4,150.60, and similar to the one observed for Cluster 2 (\$10,328.50 $\pm$ \$4,923.47). Clusters 3 and 5 have statistically equal average CDP/c values, i.e., \$44,846.74 $\pm$ \$25,956.59 and \$43,524.00 $\pm$ \$19,122.98 respectively.

\subsection{Analysis of correlations between SDG}
In this section, we analyse the matrix of correlations between goals to investigate the feasibility of simultaneous advances in all of them. 

If two goals are positively correlated, it might be that one facilitates the other or similar conditions promote the change in both of them. If there is a negative correlation, then one hinders the other, or it might be that the conditions that favor one limit the other one. The goals would have independent conditions if their correlation is close to zero. 

\begin{figure}[ht]
\centering
\includegraphics [width=1\linewidth]{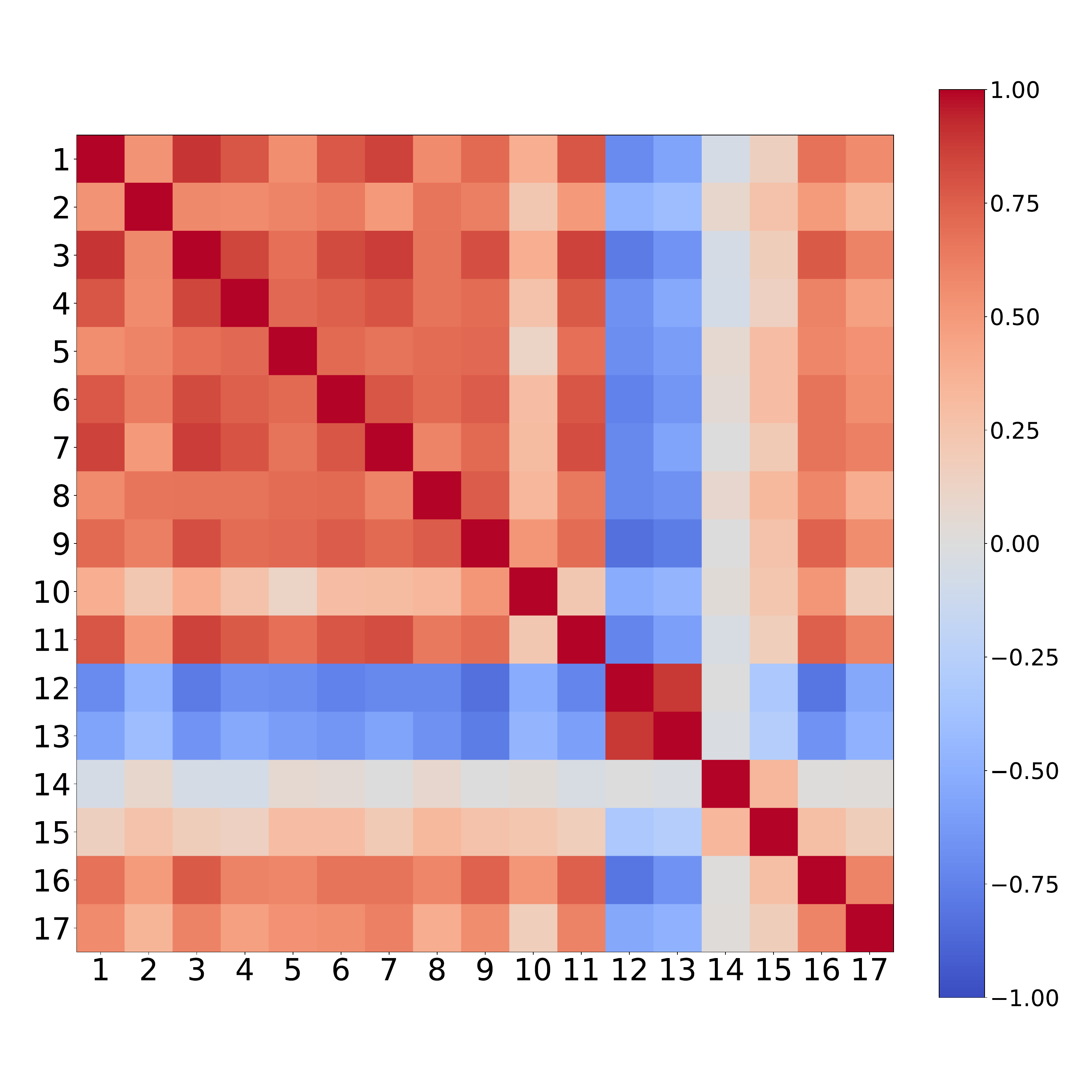}
\caption{{\bf Correlation matrix between SDG for all countries.}\\ The numbers correspond to the usual convention. Reddish tones are positive correlations and bluish ones are negative correlations.}
\label{corr}
\end{figure}

Figure~\ref{corr} presents the correlation matrix between SDGs for all countries. As expected, the auto-correlation values are at the maximum of 1. Positive correlations are depicted in red, while negative correlations are shown in blue. Notably, Goal 14 (life below water) and Goal 15 (life on land) exhibit minimal correlation with the other goals, standing out as relatively independent. Conversely, Goals 1 through 11 demonstrate positive correlations among themselves, as well as with Goal 16 (peace, justice, and strong institutions) and Goal 17 (partnerships for the goals). In contrast, Goals 12 (responsible consumption and production) and 13 (climate action) are highly positively correlated with each other, yet strongly anti-correlated with nearly all the other goals.

This observation is significant. It suggests that measures and behavioral changes aimed at reducing superfluous production and consumption could contribute to mitigating climate change but may simultaneously hinder progress towards the other SDGs.
The fact that these goals are anti-correlated with most other goals could depend on various economic factors like how the production is structured to use the resources of the planet. To gain a deeper understanding of this, we would need more detailed data on the economic structure of countries; therefore, this is beyond the present study based on the SDGs only. 
This finding underscores the complexity of sustainability as a system, where the elements interact in a nonlinear fashion, and the overall behavior cannot be understood as a simple summation of individual effects.

There are different studies on sustainability from a complexity perspective, particularly using data from China~\cite{ZHANG2022977}. These studies classify the SDGs into three categories: essential needs (Goals 2, 6, 7, 14, and 15), which represent the minimum inputs; maximum realizations (Goals 1, 3, 4, 5, 8, 10, and 16); and governance (Goals 9, 11, 12, 13, and 17), which requires compromise in competition. Governance is clearly a policy issue and inherently involves interdisciplinarity.

The fact is that our consumption and production patterns directly impact climate change, which in turn has a significant effect on the oceans~\cite{McCulloch2024}. While our findings may be qualitatively anticipated, presenting them with quantitative evidence strengthens the argument for shaping future policies. Although our goal is not to prescribe specific solutions, it is important to highlight that there are notable examples of changing our ways of living, consuming, and producing. Particularly successful models have been developed in Denmark and other regions~\cite{Maegaard2004}. Another promising example is sustainable tourism, which is not always synonymous with ecotourism~\cite{Wall1997}.
\label{corr2}

A positive correlation is evident among the first eleven goals across most clusters, with the notable exception of Cluster 5. Similarly, negative correlations are observed between the twelfth and thirteenth goals, again with Cluster 5 as the exception. Cluster 4 stands out as the only cluster exhibiting a significant presence of strong positive correlations among more than two goals, followed by Cluster 2.

\subsection{Cluster dynamics}

\begin{figure}[ht]
\centering
\includegraphics[width=1\linewidth]{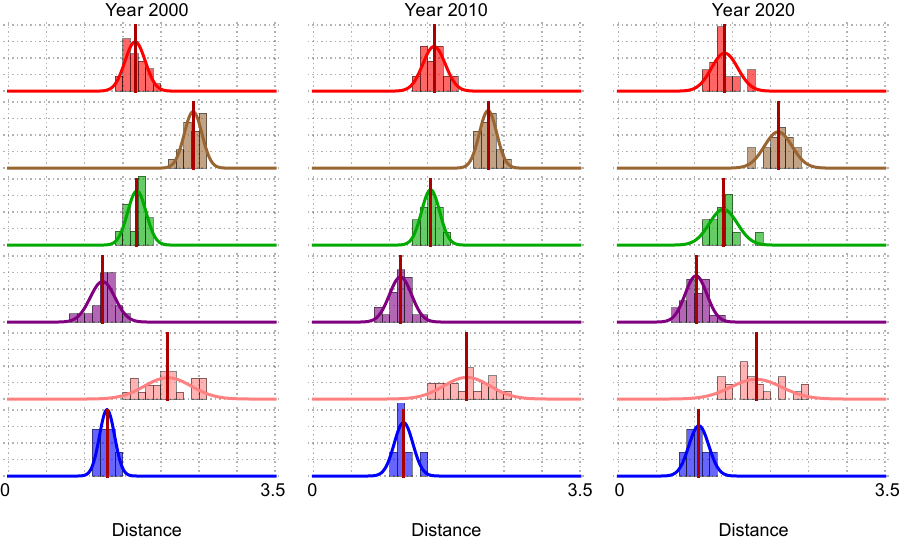}
\caption{ {\bf Fitted probability density functions.}\\Fitted probability density functions for the distribution of distances to the ideal point for three different years and for each cluster on S1 Table 3.1 (in {\bf{Supplementary material S1}} section). The colors for the clusters are the same as in Figure~\ref{fig:Map}. The Gaussian like distributions observed  indicate a central tendency and homogeneity within each cluster with respect to their progress toward the SDGs.}
\label{fig:histo}
\end{figure}

The evolution of the SDGs of each country can be tracked over time using the full 17-dimensional vector of indicators. Each component of this vector has an indicator valued between 0 and 1, representing the degree 
to which the country has achieved a specific goal. This makes the vector point
formed by these 17 values, denoted as $\vec{r}_i$, a unique vector that represents the ideal state that is being pursued. Consequently, for any given country at any point in time, the Euclidean distance to $\vec{r}_i$ can be monitored and visualized. Furthermore, by using 
the clusters described in S1 Table 2.1 (in {\bf{Supplementary material S1}} section), it is possible to calculate 
the distribution of distances to the ideal point for each year. In Figure~\ref{fig:histo} we show 
the probability density functions for the six clusters in the years 2000, 2010, and 2020. This figure reveals a continuous shift in the distribution functions which maintain 
approximately a Gaussian shape.

The Gaussian like distribution for the distances to the ideal goals within each cluster as seen in Figure ~\ref{fig:histo} suggests that the countries grouped together share a similar pattern in their progress towards the SDGs. Moreover, the fact that these clusters are Gaussian indicates that within each of them, the countries are relatively homogeneous in terms of their distance to the ideal SDG foals. This homogeneity implies that the factors influence their progress are consistent for 
the countries in each cluster and that naturally group together based on shared challenges or opportunities to reach 
the SDGs. These clusters likely reflect countries that, despite their geographical or cultural differences, face similar systemic issues or have adopted similar strategies in their progress towards the SDGs.

\begin{figure}[ht!]
\centering
\includegraphics[width=1\linewidth]{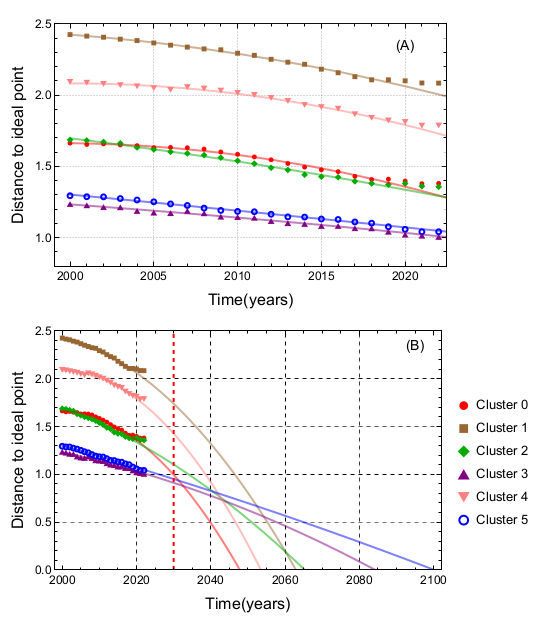}

\caption{{\bf Trajectories of clusters.}\\(A) Trajectories of clusters obtained in S1 Table 3.1 (in {\bf{Supplementary material S1}} section) toward the ideal achievement of the Sustainable Development Goals (SDGs) over time. Each line represents the projected average distance to the ideal point for a cluster of countries, with actual data points marked by dots, tracking progress from 2000 to 2022. (B) Extrapolation of same lines up to year 2100. The vertical dotted line at 2030 marks the point by which the countries of each cluster were expected to meet their SDG targets.}
\label{fig:trajectories}
\end{figure}

The mean displacement as a function of time for each cluster was calculated 
and its trajectory was fitted to the second order equation of time, as depicted 
in Figure~\ref{fig:trajectories}. This analysis 
demonstrates that the clustering procedure effectively captures information on how the clusters of countries progress 
towards the ideal target point at 2030. 
S1 Table 2.1 (in {\bf{Supplementary material S1}} section) provides the actual coefficients for the clusters identified in S1 Table 3.1 (in {\bf{Supplementary material S1}} section). Although any extrapolation based on these equations carries significant risk, it is possible to estimate the year by which, on average, the clusters will reach the ideal point. These estimates 
are included in the fifth column of S1 Table 3.1 (in {\bf{Supplementary material S1}} section). The results of extrapolation in Figure~\ref{fig:trajectories}B indicate that the trajectories of all clusters extend well beyond the 2030 target point. 

\section{Discussion}
In this study, we have shown that globally the SDGs show strong correlations. In most cases, these correlations are positive, which is considered 
beneficial. In reference~\cite{su12208729}, where the PCA was also used based on global data from 2000-2016, the authors found that synergies prevail among the SDGs, with SDG 10 (reducing inequality) being an exception. According to the results presented here, for the time span from 2000 to 2022, SDG 12 (responsible consumption and production) and SDG 13 (climate action) are strongly positively correlated with each other, but strongly negatively correlated with all the other SDGs. This finding raises a potential red flag, indicating that the current global economic system may be working against the climate action goals. This inference is based on the correlation matrix derived from the goals of each country, suggesting that the vast network of global consumption and production is a contributing factor.

There is a general agreement with reference \cite{Pradhan2017} where they found that SDG 1 (No poverty) has synergetic relationship with most of the other goals, while SDG 12 is the goal commonly associated with trade-oﬀs. Reference \cite{Bennich2023} suggest that further research is needed to clarify the systemic roles of some SDGs (e.g., SDGs 12) to understand causal relationships.
However, it is important to mention that it is difficult to compare exactly with previous global results, because the number of countries, the time span considered and the geographical regions \cite{Kostetckaia2022} vary.

It is important to note that these correlations are global. Ideally, all correlations would be positive or close to zero, indicating that actions taken toward one goal by a country do not adversely affect the others. Achieving this would require  
both internal efforts by individual countries and external efforts through cooperation among countries in terms of economic, political, and other relations. However, each country or group of countries may exhibit different correlations, as we have observed. As mentioned earlier, globally the SDG 12 and SDG 13 are strongly positively correlated with each other but are strongly negatively correlated with all the other SDGs. However, groups 3 and 5 show scores above the average except for the SDGs 12 and 13, while groups 1 and 4 exhibit an inverse behavior. Groups 0 and 2 do not present these correlations.

In reference~\cite{LAMICHHANE2021125040}, the PCA is also used, but only, 
for 35 countries of the OECD. The authors found that some countries perform significantly better compared to others, reflecting the heterogeneity within the OECD group, which is based on political agreements. The advantage of our study lies in the combined use of PCA and t-SNE, which leads to an original clustering, based on real data. It is important to notice that clusters were obtained using solely SDG scores, showing that groups of countries having akin SDG pattens exhibit similar behavior, primarily due to geographical, cultural, and socio-economic features.

Based on our results it is possible to provide some general insights. African and Asian countires (Clusters 1 and 4), which are the ones having the lowest GDP/c, \$1,464.44 and \$3,941.89, respectively, show low SDG scores for Goals 1 to 11, 16 and 17 but posses the best performance with respect to responsible consumption and production and 
climate action (Goals 12 and 13); in contrast, the developed countries (Clusters 3 and 5), having the highest GDP/c, \$44,846.74 and \$43,524.00, respectively, show good achievements for Goals 1 to 11, 16 and 17 but have problems to improve their scores for Goals 12 and 13. The positive correlation between Goals 12 and 13 is very clear here, responsible consumption and production helps prevent climate change, but some other the economic factors seem to act against these goals. Latin America (Cluster 2), having the GDP/c \$10,328.50, has a different SDG pattern, showing the SDG scores above the mean for Goals 1 to 9, 11 to 14 and 17, and the lowest score for Goal 10, i.e., related to actions to reduce inequalities. 


None of the 107 countries we studied has all SDGs with either uniformly good or bad marks. They all display different standardized score patterns that are clearly visible in Figure~\ref{fig:ClutSNE}. For instance, group 4 (Asian countries) exhibits a pattern that is almost the inverse of that observed in the so-called Western countries. However, similar countries, such as the USA and Canada, belonging to groups five and three respectively, differ in their environmental policies regarding the SDGs 14 and 15.

In Figure~\ref{fig:trajectories}, the trajectories of the mean distance to an ideal point for each cluster from S1 Table 3.1 (in {\bf{Supplementary material S1}} section) are shown. All clusters are pursuing an ideal point, starting from different initial conditions. It is important to note that in 2020, the progress slowed down, most likely due to the COVID-19 pandemic.

There seems to be a relation among the standardized scores patterns and the dynamics.   
There are different rates of change. It appears that the wealthiest countries, in general, are the slowest to achieve the overall goal, while the less wealthy countries make faster progress. In any case, 
extrapolating the current tendencies (see Appendix C), no country would be reaching
the goals by 2030.

The patterns found in Fig.~\ref{fig:ClutSNE} 
 could be interpreted as an imprint from some common policies implemented in the members of a cluster, that produce close results. It 
does not seem to be a coincidence that clusters reveal those patterns related to
different trajectories in Fig.~\ref{fig:trajectories} that show how a set of countries 
progresses in reaching the goals. Although the trajectories shown in the plot suggest that some clusters of countries may be on a path toward achieving the Sustainable Development Goals (SDGs) a few decades after the 2030 target, this optimistic outlook is likely to be disrupted by several significant global challenges. The main ones are the 1.5°C and 2°C global warming thresholds expected around 2030 and 2050, respectively~\cite{ipcc2018}. Crossing these thresholds could exacerbate climate-related disasters, undermine ecosystems, and severely impact agricultural and water resources, making the achievement of the SDGs much more difficult. 

Additionally, the projected population growth to 9 billion by 2037 and 10 billion by 2055 will place increasing pressure on natural resources~\cite{unpop2019}, which would intensify the  
competition and lead potentially 
to the depletion of critical resources such as freshwater and minerals by mid-century. The biodiversity crisis projected around 2050, if the current trends continue, could further delay progress by destabilizing the ecosystems which  
many human activities depend on~\cite{cbd2020}. Economic shifts, technological advances, and the persistent effects of the COVID-19 pandemic~\cite{Monti2021} also introduce uncertainties that could alter the course of sustainable development. Therefore, while the plotted extrapolated trajectories indicate a much delayed  potential progress, the convergence toward the SDGs even decades later could 
be challenged by these  
obstacles, casting doubt on the feasibility of achieving these goals even then. 

The method proposed to reduce variables and classify countries by their
characteristics in the 17 dimensional space 
allows the identification of particular changes associated with 
a representative subset of countries. At the same time, an analysis like the one presented in Figure~\ref{fig:trajectories}, gives the possibility to identify and quantify the impact of global phenomena, like a pandemic, on the data.

This should be concerning. If the global inertia of twenty years stopped for several regions, this could be interpreted as poorer countries being less resilient to crises to achieve sustainability goals. Thus, unexpected future crises could very well hinder the progress towards sustainability.

\section{Concluding Remarks}

To summarize our study we have performed a comprehensive multi-phase analyses by using PCA, t-SNE, and DBSCAN clustering, as well as parallel plot methodologies to unravel a detailed landscape of SDGs. The depth of the present set of methods enabled us to underpin the criticality of a segmented approach, addressing both the global and region-centric challenges. Through this type of analytical rigor, we wanted to demonstrate that the policymakers and researchers could be equipped with a granular roadmap, emphasizing collaborative initiatives, data-informed strategies, and sustainable advancement.

From our study, two main conclusions can be drawn: 

i) the available data could give detailed descriptions at the medium scale, for countries to organise their response to the 2030 Agenda by using common policies while paying attention to 
their own geographical and socioeconomic situation, but with similar inputs and outputs as others. These regions can be identified and described accurately by using machine learning methods.

ii) Global crises leave footprints in the data that can be identified and analysed. For example, in some identified regions, Covid-19 turned out to brake the slow progress or inertia before the pandemic and then continued to develop in opposite direction, i.e., in trajectories away from achieving the SDGs. This development would make Agenda 2030 unattainable.  

There could be several reasons for differences in development trajectories. One of the main ones could be that different countries strive for different goals due to their regional, cultural, historical, social, and economic differences. This is seen by the clustering being closely correlated with geographical regions. Also we expect that the climate change plays a role globally, but the ways different countries suffer and deal with it has local characteristics. Furthermore, it has turned out that actions to achieve a certain goal may jeopardize the advance on other goals. This is because SDGs have turned out to interact both in positive and negative ways. For instance, the SDG 12 (responsible consumption and production) is particularly polemic, because it is strongly related with cultural aspects and competes with the SDG 13 {climate action}.
 
 In the light of these observations, the relationships between different SDGs are very complex. Given that as even the most optimistic projections show that clusters of countries are far from achieving the ideal SDGs, there is a need for accelerated action. This might include a re-evaluation of the strategies being used to pursue the SDGs. Furthermore, it is evident that the individual approaches currently being used to achieve the SDGs are not working well for most countries. Thus a new systemic approach may be necessary to achieve these goals. This could involve creating new SDGs that depend not just on individual countries but on a diverse group of countries working together to form a support network. 

\section*{Acknowledgments}

A. García acknowledges financial support from the Center of Complexity Sciences C3 at UNAM. M. Nunez acknowledges engaging conversations with Dr. Vera Chiodi from La Sorbonne.

\newpage

\nolinenumbers


\section*{Supplementary material}

\counterwithin{figure}{section}
\setcounter{figure}{0} 
 
\section {Definition of SDG}
\label{sec:SDG}
In here we show in a familiar graphical way the description of the SDG's to facilitate their identification if needed.

\begin{figure}[ht]
\centering
\includegraphics[width=1\linewidth]{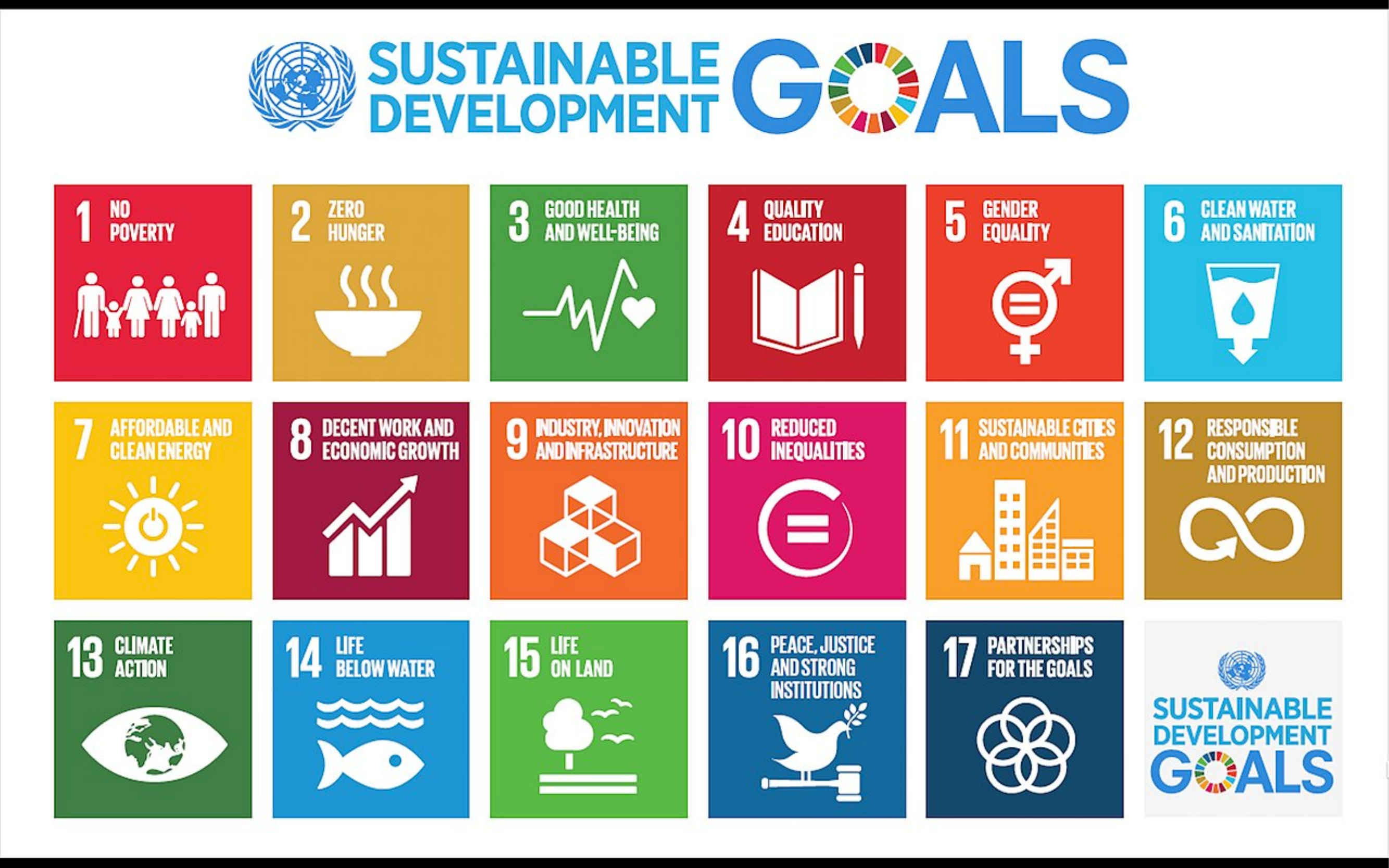}
\caption{Meaning of the 17 SDG according to the UN }
\label{SDG}
\end{figure}
\vspace{1in}
\counterwithin{table}{section}

\section{Countries in the different clusters}

We show explicitly the names of the countries belonging to each cluster in S1 Table \ref{tablaclusters}. Notice that cluster Number -1 is not really a cluster, 
it groups dots classified by DBSCAN as not belonging to any cluster since they lie in low density zone. Nevertheless, we include them for completeness. Also notice that few countries are in more that one cluster. This means that during the time span of the analysis they changed their situation 

\begin{table}
\centering
\caption{Countries grouped into clusters based on their SDG trajectories. The data underwent a sequential analysis: first, Principal Component Analysis (PCA) was utilized for dimensionality reduction. Then, t-Distributed Stochastic Neighbor Embedding (t-SNE) was applied to further map the data into a two-dimensional space, Finally, the DBSCAN clustering algorithm was used to group countries based on the resulting patterns.}

\label{tablaclusters}
\begin{tabularx}{\textwidth}{|c|X|}
\hline
\textbf{Cluster} & \textbf{Countries} \\
\hline
-1 & Cyprus, United Arab Emirates, parts of the trajectories of: Bangladesh, Benin, Malta \\
\hline
0 & Africa: Algeria, Cabo Verde, Egypt (Arab Rep.), Gabon, Morocco, Sao Tome and Principe, Senegal, Tunisia; America: El Salvador, Panama, Uruguay ; Asia: Georgia, Iran (Islamic Rep.), Jordan, Lebanon, Malaysia,  Maldives, Russian Federation, Thailand, Türkiye; Europe:  Albania, Bosnia and Herzegovina, Bulgaria, Montenegro; Oceania:  Fiji \\
\hline
1 & Africa: Angola, Benin, Cameroon, Congo (Dem. Rep.), Congo (Rep.), Cote d'Ivoire, Gambia (The), Guinea, Kenya, Liberia, Madagascar, Mozambique, Nigeria, Senegal, Sierra Leone, Tanzania, Togo; America: Haiti; Oceania: Papua New Guinea\\
\hline
2 & America: Argentina, Brazil, Chile, Colombia, Costa Rica, Dominican Republic, Ecuador, Jamaica, Mexico, Peru, Uruguay, Venezuela (RB); Asia: Philippines \\
\hline
3 & America: Canada; Europe: Belgium, Bulgaria, Croatia, Denmark, Estonia, Finland, Germany, Greece, Iceland, Ireland, Italy, Latvia, Lithuania, Malta, Netherlands, Norway, Poland, Romania, Slovenia, Sweden, United Kingdom; Oceania: Australia \\
\hline
4 & Afria: Comoros, Djibouti,  Gabon, Ghana, Mauritania, Mauritius, Namibia, Sao Tome and Principe, Somalia, South Africa, Sudan; America: Guatemala, Haiti, Honduras, Nicaragua; Asia: Bangladesh, China, India, Indonesia, Myanmar, Pakistan, Romania, Sri Lanka, Vietnam, Yemen (Rep.) \\
\hline
5 & America: United States; Asia: Israel, Japan, Korea (Rep.); Europe: France, Portugal, Spain\\
\hline 
\end{tabularx}
\end{table}

The 10 countries which changed their classification: Bangladesh from ‘-1’ (2000-2017) to cluster 4 (2018.2022); Benin from cluster 1 (2000-2017) to ‘-1’ (2018-2022); Bulgaria from cluster 0 (2000-2006) to cluster 3 (2007-2022); Gabón from cluster 4 (2000-2001) to cluster 0 (2002-2022); Haiti from cluster 4 (2000-2020) to cluster 1 (2021-2022); Malta from ‘-1’ (2000-2010) to cluster 3 (2011-2022); Romania from cluster 4 (2000-2006) to cluster 3 (2007-2022); Sao Tome and Principe from cluster 4 (2000-2005) to cluster 0 (2006-2022); Senegal from cluster 1 (2000-2018) to cluster 0 (2019-2022); Uruguay from cluster 0 (2000-2004) to culster 2 (2005-2022).

\section {Dynamics}
\label{dynamics}
In here we show the  values of the coefficients of the second order
equation used to fit the dynamical behavior of the goals. According to this the prediction of the dates to achieve the ideal score is also  shown in S1 Table
\ref{tab:extrapol}.

\begin{table}[ht!]
    \centering
     \begin{tabular}{|ccccc|}
\hline
Cluster& $a$ & $b$ & $c$ &  year to zero \\ \hline
0 & -2799.59 & 2.80247 & -0.000700922 & 2048 \\
1 &  -1881.52 & 1.89254 & -0.000475281 & 2063 \\
2 &  -669.863 & 0.686105 & -0.000175162 & 2066 \\
3 & -269.743 & 0.279532 & -0.000072022 & 2085 \\
4 & -2903.6 & 2.90572 & -0.000726443 & 2054 \\
5 &  -49.7872 & 0.0622445 & -0.00001835 & 2101 \\
  \hline
     \end{tabular}
     \caption{Best fitting coefficients of equation $r(t)=a +b t+ c t^2$ for each cluster of countries obtained in S1 Table~\ref{tablaclusters}. The last column is the extrapolated time to plenty achieve all the goals. Fitting was done without 2020, 2021 and 2022 data, to suppress the effect of COVID pandemic. }
     \label{tab:extrapol}
 \end{table}

\section{The t-SNE dimensional reduction method.}
\label{sec:tsne}

The t-Distributed Stochastic Neighbor Embedding algorithm is a non-linear dimensionality reduction technique designed to visualize high-dimensional data by mapping it into a lower-dimensional space . Its main objective is to preserve the local structure of the data by ensuring that points that are close in the high-dimensional space remain close in the lower-dimensional representation.

\subsection{ Probability Distributions in High Dimensions}
\label{subsec:perplexity}

t-SNE starts by converting the pairwise Euclidean distances between points in the high-dimensional space into conditional probabilities. These probabilities represent the similarity between points. Specifically:
\[
P_{j|i} = \frac{\exp\left(-\|x_i - x_j\|^2 / 2\sigma_i^2\right)}{\sum_{k \neq i} \exp\left(-\|x_i - x_k\|^2 / 2\sigma_i^2\right)}
\]
Here, \(P_{j|i}\) is the probability that point \(x_i\) picks \(x_j\) as its neighbor, assuming a Gaussian distribution centered at \(x_i\) with variance \(\sigma_i^2\).

\subsubsection{Perplexity}
Perplexity is a crucial parameter in t-SNE that controls the balance between local and global aspects of the data structure. It can be interpreted as a smooth measure of the effective number of neighbors for each point. The perplexity is defined as:
\[
\text{Perplexity}(P_i) = 2^{H(P_i)}
\]
where \(H(P_i)\) is the Shannon entropy of the conditional probability distribution \(P_i\):
\[
H(P_i) = -\sum_{j} P_{j|i} \log_2 P_{j|i}
\]
Perplexity is typically chosen by the user and affects the value of \(\sigma_i\) for each data point, thereby controlling how the local neighborhood is defined.

\subsection{ Mapping to Lower Dimensions}

In the lower-dimensional space, t-SNE aims to find a similar probability distribution \(Q_{ij}\) that reflects the similarities between points:
\[
Q_{ij} = \frac{\left(1 + \|y_i - y_j\|^2\right)^{-1}}{\sum_{k \neq l} \left(1 + \|y_k - y_l\|^2\right)^{-1}}
\]
Notice that in the lower-dimensional space, t-SNE uses a Student's t-distribution with one degree of freedom (essentially a Cauchy distribution) instead of a Gaussian distribution. This choice allows t-SNE to handle the so-called "crowding problem," ensuring that distant points are placed further apart.

\subsection{ Optimization Objective}

The t-SNE algorithm seeks to minimize the Kullback-Leibler (KL) divergence between the probability distributions \(P_{ij}\) (in the original space) and \(Q_{ij}\) (in the lower-dimensional space):
\[
KL(P \| Q) = \sum_{i \neq j} P_{ij} \log \frac{P_{ij}}{Q_{ij}}
\]
The optimization adjusts the positions of the points in the lower-dimensional space to minimize this divergence, resulting in a configuration where similar points (high \(P_{ij}\)) are close together, and dissimilar points are far apart.

\end{document}